\documentclass[10pt, conference, compsocconf]{IEEEtran}
\IEEEoverridecommandlockouts
\usepackage{cite}
\usepackage{amsmath,amssymb,amsfonts}
\usepackage{algorithmic}
\usepackage{graphicx}
\usepackage{textcomp}
\usepackage{xcolor}
\def\BibTeX{{\rm B\kern-.05em{\sc i\kern-.025em b}\kern-.08em
    T\kern-.1667em\lower.7ex\hbox{E}\kern-.125emX}}

\usepackage{tikz}
\usepackage{mathtools}
\usepackage{amsfonts}

\usepackage{graphicx}
\usepackage{enumerate}

\usepackage[shortlabels]{enumitem}
\usepackage{subfigure}
\usepackage{caption}
\usepackage{dsfont}

\usepackage{xcolor,colortbl}
\usepackage{hyperref}
\definecolor{Gray}{gray}{0.85}

\usepackage{booktabs}
\usepackage{stackengine}

\begin{document}

\title{\LARGE \bf A Novel Graphical Lasso based approach towards Segmentation Analysis in Energy Game-Theoretic Frameworks
}

\author{Hari Prasanna Das\textsuperscript{\textdagger}, Ioannis C. Konstantakopoulos\textsuperscript{\textdagger}, Aummul Baneen Manasawala\textsuperscript{\ddag},\\ Tanya Veeravalli\textsuperscript{\textdagger}, Huihan Liu\textsuperscript{\textdagger} and Costas J. Spanos\textsuperscript{\textdagger}\\
\textsuperscript{\textdagger}Department of Electrical Engineering and Computer Sciences, University of California, Berkeley\\
\textsuperscript{\ddag}Department of Industrial Engineering and Operations Research, University of California, Berkeley\\
Emails: {\{\tt hpdas,ioanniskon,abf,tveeravalli19,liuhh,spanos\}}@berkeley.edu
}

\maketitle

%%%%%%%%%%%%%%%%%%%%%%%%%%%%%%%%%%%%%%%%%%%%%%%%%%%%%%%%%%%%%%%%%%%%%%%%%%%%%%%%
\begin{abstract}
%energy game theoretic frameworks have been prevalent
%incentive design process is conventional
%key to incentive design is better understanding of features leading to human decision making and segmentation
%we propose a graphical lasso based segmentation mechanism
%we improve explainability of the model using grangers causality
%proposed segmentation....
%present possible avenues of future research
Energy game-theoretic frameworks have emerged to be a successful strategy to encourage energy efficient behavior in large scale by leveraging human-in-the-loop strategy. A number of such frameworks have been introduced over the years which formulate the energy saving process as a competitive game with appropriate incentives for energy efficient players. However, prior works involve an incentive design mechanism which is dependent on knowledge of utility functions for all the players in the game, which is hard to compute especially when the number of players is high, common in energy game-theoretic frameworks. Our research proposes that the utilities of players in such a framework can be grouped together to a relatively small number of clusters, and the clusters can then be targeted with tailored incentives. The key to above segmentation analysis is to learn the features leading to human decision making towards energy usage in competitive environments. We propose a novel graphical lasso based approach to perform such segmentation, by studying the feature correlations in a real-world energy social game dataset. To further improve the explainability of the model, we perform causality study using grangers causality. Proposed segmentation analysis results in characteristic clusters demonstrating different energy usage behaviors. We also present avenues to implement intelligent incentive design using proposed segmentation method.
\end{abstract} 
\begin{IEEEkeywords}
Segmentation Analysis, Energy Game-Theoretic Frameworks, Graphical Lasso, Smart Building
\end{IEEEkeywords}

%%%%%%%%%%%%%%%%%%%%%%%%%%%%%%%%%%%%%%%%%%%%%%%%%%%%%%%%%%%%%%%%%%%%%%%%%%%%%%%%
\section{Introduction}
Energy consumption of buildings, both residential and commercial, account for approximately 40\% of all energy usage in the U.S.~\cite{mcquade2009}. In efforts to improve energy efficiency in buildings, researchers and industry leaders have attempted to implement control and automation approaches alongside techniques like incentive design and price adjustment to more effectively regulate the energy usage~\cite{Konstantakopoulos:EECS-2018-139}. The heterogeneity of user preferences in regard to building utilities  is  considerable  in  variety  and  necessitates  a  system that can adequately account for differences from one occupant to  another.  With  this  in  mind,  focus  has  shifted  towards modeling  occupant  behavior  to  incorporate  their  preferences in building control and automation ~\cite{delzendeh2017the}. The behavior models can then be studied to introduce initiatives to encourage energy efficient behaviors among the occupants/energy users. Furthermore, the occupants of a building typically lack the independent motivation necessary to contribute to and play a key role in the control of smart building infrastructure. A building manager, acting as the connection between energy utilities and the end users, can encourage participation and energy-efficient behavior among occupants in many ways. One of the successful methods proposed and improved over the years is a game-theoretic framework, which creates a
friendly competition between occupants/users, motivating them to individually consider their own energy usage and hopefully, seek to improve it to have a better score/achieve a lucrative incentive in the game \cite{konstantakopoulos2019deep}.

 Energy game-theoretic frameworks have proven to be capable of extracting the occupant behavior and possibly impacting a change in them by engaging the users in the process of energy management, integrating seamlessly through the use of cyber-physical technology by leveraging humans-in-the-loop strategy \cite{konstantakopoulos2019deep,zou2019consensus,Liu_thermal_comfort}. Such game-theoretic frameworks can be thought of as a sensor-actuator system. Through their participation in the game (the sensor), the behavior of users is observed, which then is treated as the input to an incentive design process (the actuator). The incentives offered can motivate users to improve upon their energy usage behaviors to achieve better energy efficiency, signifying the importance of an intelligent incentive design in success of such frameworks. Although all such frameworks  aim to achieve a long term or permanent improvement in the energy usage behaviors among the users, the aim is seldom achieved after the completion of energy game, mostly attributed to the lack of an intelligent and adaptive incentive design process. The incentive design process in prior works is dependant on utility functions of every player in the game, which is hard to compute as energy game-theoretic frameworks involve participation of a large number of energy users, and hence often approximated. Instead, the utility/energy usage behavior of the players in such a large scale frameworks can be segmented into a relatively small number of clusters, and incentives can be designed to tailor each cluster assuming players in a cluster behave synchronously. Energy utility companies frequently use such segmentation techniques for optimal planning of demand response, load shedding, and microgrid applications~\cite{energy_utility}. We consider the design of a smarter segmentation analysis as a solution for intelligent incentive design for energy game-theoretic frameworks as the objective of this research work.

The above goal can be achieved by learning the factors leading to human decision-making, and using the knowledge to devise a novel agent segmentation method. The segmentation analysis in an energy game-theoretic framework with high dimensional data requires powerful yet computationally efficient statistical methods. A possible candidate, Graphical Lasso algorithm ~\cite{mainbook}, has been widely applied on different scientific studies due to its sparsity property ($\ell_1$ penalty term) and efficiency \cite{glasso_biology,cvpr_scene}. The potential of Graphical Lasso can be innovatively combined with player segmentation. Towards this, we enable new avenues by combining both concepts and applying it on a social game data set\cite{konstantakopoulos2019deep} to classify the energy efficiency behaviors among building occupants. We explore the causal relationship between different features of the agents using a versatile tool, Grangers Causality~\cite{granger1980testing}, which leads to a deep understanding of decision-making patterns and helps in integrating explainable game theory models with adaptive control or online incentive design. With the advent of explainable Artificial Intelligence (AI), there has been a massive move towards making statistical models explainable. Thus we propose an explainable, rather than just a black box model. To summarize, our contributions are threefold:
\begin{itemize}[leftmargin=*]
    \item Novel segmentation analysis using an explainable statistical model at the core towards learning agent's (building occupants) features characterizing their decision-making in competitive environments.
    \item Characterization of causal relationship among several contributed features explaining decision-making patterns in agent's actions.
    \item Introduction of possible avenues which can prove beneficial in improving building energy efficiency by using the proposed segmentation analysis method 
\end{itemize}

\section{Related Work}\label{sec:related_work}
Energy game theoretic frameworks have enabled an effective platform for incorporating energy efficient behavior among the building occupants in a smart building. They involve the important aspect of human participation in building control, otherwise lacking in many conventional modelling approaches including passive Hidden Markov Models (HMMs)~\cite{ai2014occupancy}. A number of such frameworks have been introduced over the years ~\cite{konstantakopoulos2019deep,knol:2011aa,Dahleh2010smartCom}, which have proven to have significant energy conservation during post-game period as compared to pre-game period.

Just like nutritional requirements vary from one person to another, incentive designs must be tailored to be suitable for every player, to have an impact on their behaviors ~\cite{konstantakopoulos2016inverse,konstantakopoulos2017leveraging,konstantakopoulos2018robust}. For such incentive design, above game-theoretic frameworks rely on knowledge of utility functions of the players in the game, which is hard to compute in the scenario of energy game theoretic frameworks due to the complexity and scale. Authors in ~\cite{li:2014aa} propose a nash-equillibrium based approach for utility estimation. In ~\cite{ratliff2014social}, authors formulate the utility estimation problem as a convex optimization problem by using first-order necessary conditions for nash equilibria, and then create an affine map along with energy consumption to derive the utilities. All these methods are hard to scale when the number of players is high. Instead, we can segment the utilities of players into clusters by learning features characterizing human decision-making in competitive environments, and perform an incentive design for the clusters so obtained. We derive inspiration for agent segmentation owing to the fact that customer segmentation has been successfully utilized in energy systems~\cite{energy_utility}. The energy usage behavior exhibited by each player in a cluster is expected to be similar, which has statistical justification as the number of possible clusters in the data is computed using clustering algorithms.  

Towards this we use high dimensional real-world data. We use the graphical lasso algorithm as a powerful tool to understand the latent conditional dependence between variables~\cite{mainbook}. This in turn provides insights into how different features interplay among each other. Historically, Graphical Lasso has been used in various fields of science, ranging from study of how individual elements of the cell interact with each other~\cite{glasso_biology} and to the broad area of computer vision for scene labelling~\cite{cvpr_scene}. A modified version of the original algorithm, named time-varying graphical lasso, has been used on financial and automotive data~\cite{HallacPBL17}. However, the novelties of graphical lasso has not been well utilized in the area of energy cyber-physical systems. We use Granger causality\cite{granger1980testing} to explain the causal relationship between the features in energy usage behavior of agents in social game. It has been widely used in the energy domain in applications such as deducing the causal relationship between economic growth and energy consumption~\cite{chiou2008economic}.

Knitting novel segmentation algorithms and their application to energy game-theoretic frameworks together, we employ graphical lasso algorithm for customer segmentation on social game dataset and present an explainable model, helpful both in understanding inherent factors leading to energy efficiency in buildings and in intelligent incentive design.
\section{Methods}\label{sec:data_analysis}
\subsection{Energy Game-Theoretic Dataset}
The dataset used for our work is from a energy social game experiment to encourage energy efficient resource consumption in a smart residential housing, as introduced in \cite{konstantakopoulos2019deep}. Authors in ~\cite{konstantakopoulos2019deep} designed a social game among occupants of residential student housing apartments at an university campus. They make use of Internet of Things (IoT) sensors to allow the occupants to monitor their room's lighting (desk and ceiling light) and ceiling fan usage via a personal web-portal account as they participate in the energy social game for maximizing their incentives. The above game-theoretic framework is modelled under the umbrella of a multiplayer non-cooperative game. The dataset consists of per-minute time-stamped reading of each resource's (desk light, ceiling light and ceiling fan) status, accumulated resource usage
(in minutes) per day, resource baseline, gathered points (both
from game and surveys), occupant rank in the game over
time and number of occupant’s visits to the web portal. It also contains features related to time of day (morning vs. evening), time of week (weekday vs. weekend) and college
schedule feature indicators for dates related to breaks, holidays, midterm and final exam periods. Additionally, the dataset incorporates the external weather metrics during the experimental run.

\subsection{Trade-off between Supervised/Unsupervised Segmentation}
\begin{figure}[!ht]
    \includegraphics[width=0.49\textwidth]{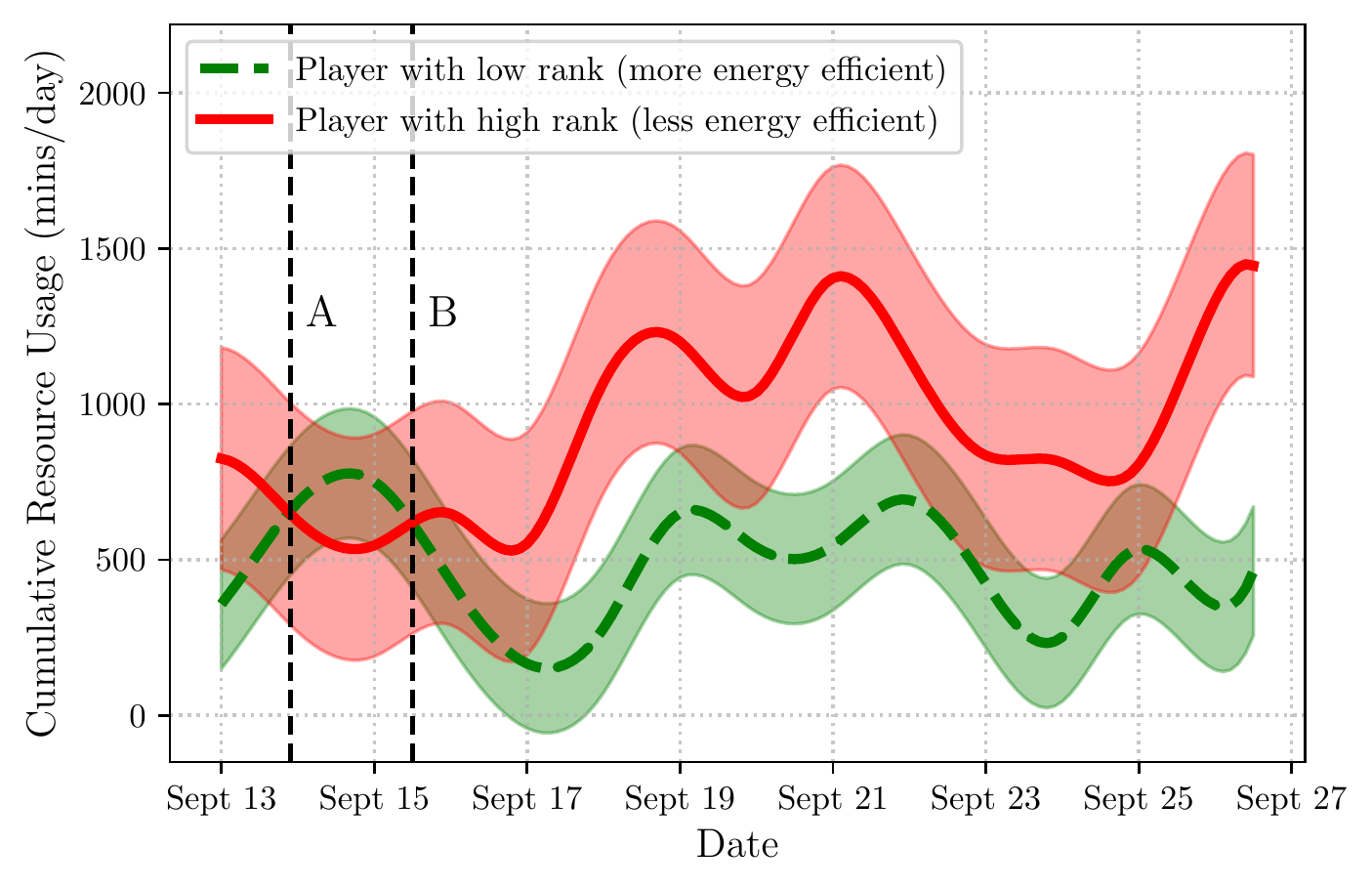}
    \caption{Variation of cumulative energy resource usage (mins/day) for a player with low rank (high energy efficient) and another with high rank (low energy efficient)}
    \label{fig:variational_energy_efficiency}
\end{figure}

For the purpose of segmentation analysis, both supervised and unsupervised segmentation methods can be implemented on the social game dataset. Supervised methods require a label to classify data with similar labels together. For the dataset in hand, the label we have is the rank of the player in the game, which in turn indicates their energy efficiency characteristics as compared to other players in the game (i.e. a player with less rank is more energy efficient). We use rank as the label to classify players into different groups. But, such a classification method groups different players together as per their overall rank, and does not take into account the distribution of their energy efficiency characteristics across different scenarios such as time. For instance, Figure~\ref{fig:variational_energy_efficiency} shows the distribution of cumulative resource usage (in minutes/day) for a player having low rank (high energy efficiency) and a player having high rank (low energy efficiency), with some curve smoothing across a duration of the game period. It can be observed that the high energy efficient player performs sub-optimally (uses more energy resources) between the times A and B than the low energy efficient player. In an ideal scenario, for every player, the data samples corresponding to low energy efficient behavior should be clustered separately than high energy efficient behaviors so as to have an accurate understanding of the interplay of features governing human decisions for energy efficiency. In this case, unsupervised clustering proves helpful and clusters together similar behaviors. But in this case, the output of unsupervised clustering is just a number of clusters with no labelling about the energy efficiency characteristics exhibited by that cluster. So, to summarize, supervised classification provides insight into an overall picture of how different classes of energy-efficient players behave, but fails to capture the distribution of behaviors. On the other hand, unsupervised clustering captures the latter accurately, but does not provide any information on labels of the clusters generated. This poses a trade-off between supervised classification and unsupervised clustering methods for application in energy game datasets. 
\subsection{Proposed Segmentation Method}\label{sec:method}
\begin{figure}[!ht]
\includegraphics[width=0.48\textwidth]{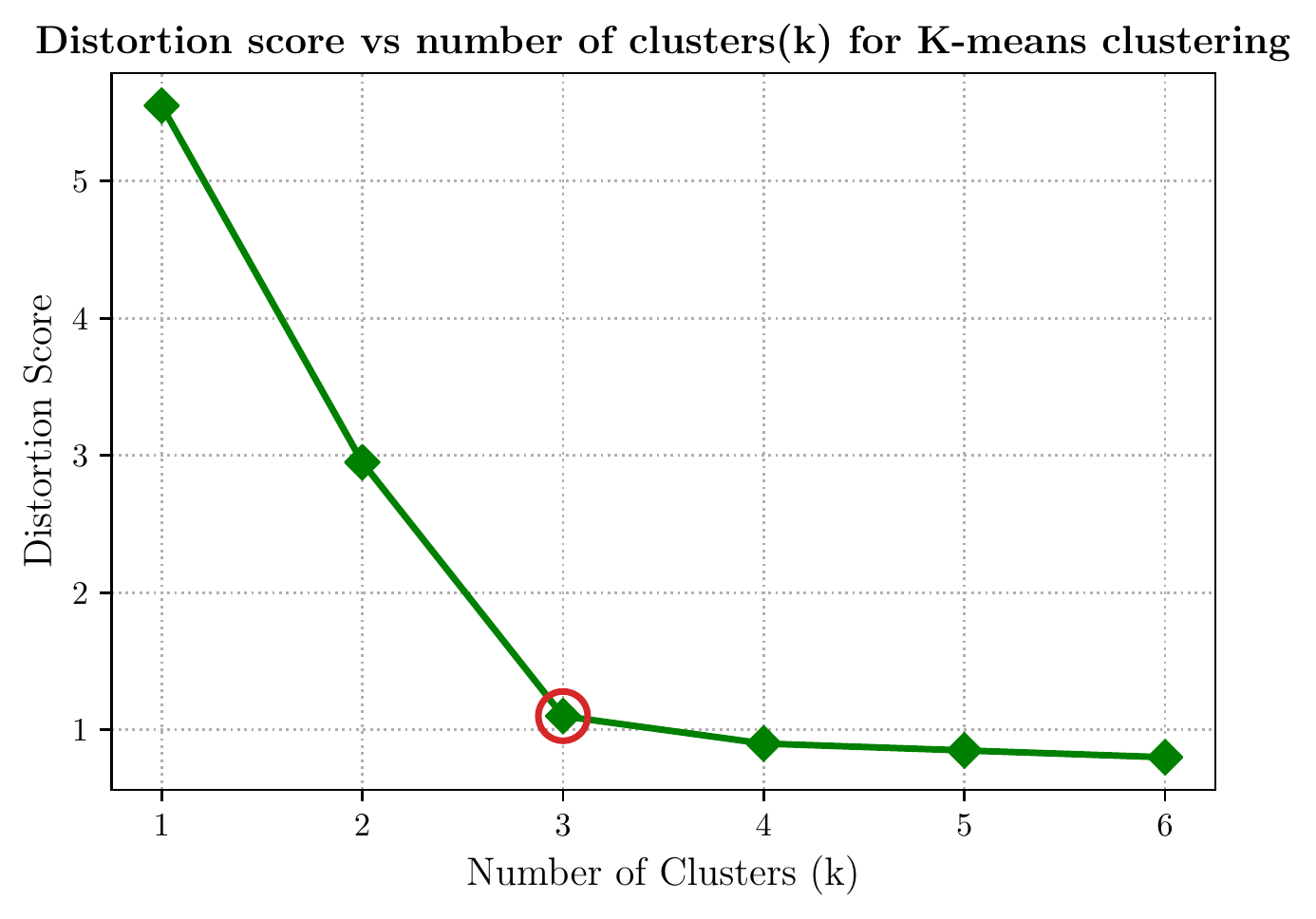}
\caption{Elbow Plot to choose optimal number of clusters}
\label{fig:elbow_plot}
\end{figure}

The trade-off mentioned in previous section signals to use the novelty of both unsupervised and supervised segmentation together to build an optimal model. Knitting together both the methods via a powerful tool, the graphical lasso algorithm, we present a novel methodology to perform segmentation in energy game-theoretic frameworks. We employ the k-means algorithm for unsupervised clustering. First, the optimal number of clusters in the dataset is derived using elbow method and silhouette scores. An elbow plot is a plot between the distortion score (a measure of closeness of data points to their assigned cluster center) vs the number of clusters. The optimal number of clusters is determined to be corresponding to drastic change in the rate of reduction in distortion score. The elbow plot for energy social game dataset, obtained in an unsupervied manner is given in Figure~\ref{fig:elbow_plot}. The optimal number of clusters is determined to be 3. 
We also use another metric, the silhouette score to confirm the optimal number of clusters. The silhouette score $\in[-1,1]$, is a measure of how similar an object is to its own cluster compared to other clusters. A high value indicates that the object is well matched to its own cluster and poorly matched to neighboring clusters. The silhouette score corresponging to each number of clusters is given in Table~\ref{tab:silhouette_scores}. Note that the score is the highest (shaded in blue) for the number of clusters as 3. 
\begin{table}[h!]
\setlength\arrayrulewidth{0.8pt}
\centering
\begin{tabular}{ |c|c|c|c|c| } 
\rowcolor{Gray}
 \hline
 No. of Clusters & 2 & 3 & 4 & 5 \\ 
 \hline
 Silhouette Scores & 0.684 & \cellcolor{blue!25}0.749 & 0.611 & 0.540 \\ 
 \hline
\end{tabular}
\caption{Silhouette Scores for different number of clusters}
\label{tab:silhouette_scores}
\end{table}
\begin{figure*}[t]
\centering
  \includegraphics[height=4cm]{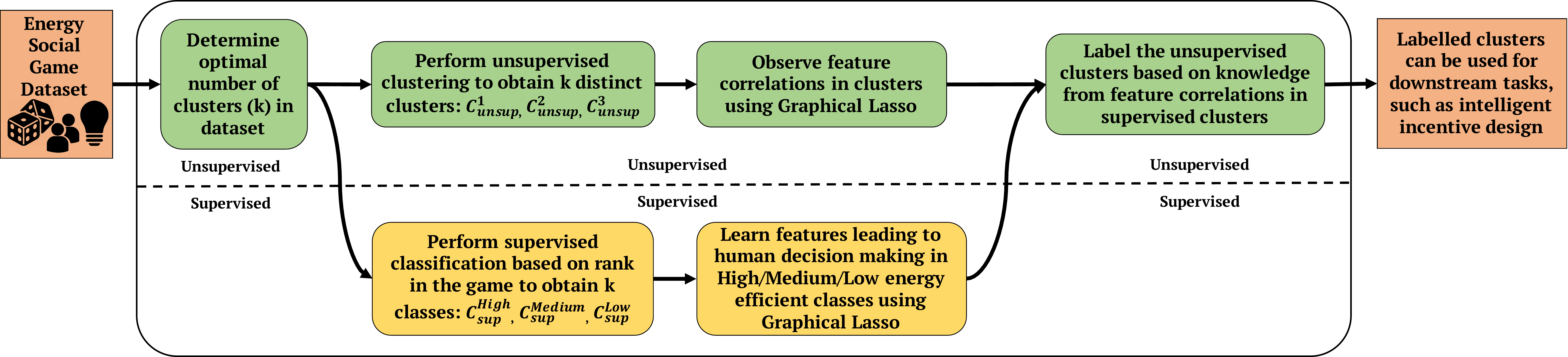}
  \caption{Overview of the proposed segmentation method}
  \label{fig:proposed_method}
\end{figure*}
Following this, we use Minibatch K-means algorithm with k= 3 (optimal number of clusters in the data) to obtain the clusters. Let the clusters obtained be represented by $C_{unsup}^{1}$, $C_{unsup}^{2}$ and $C_{unsup}^{3}$. Since the dataset correspond to energy usage behavior of the players, the three clusters so obtained correspond to high, medium and low energy efficient behaviors. We then use supervised classification and graphical lasso to label the unsupervised clusters. We divide the players into three classes in a supervised way taking the ranks of the users as the label. 

Let the players be denoted by $P_1$, $P_2$, $\cdots$, $P_m$ and the data points corresponding to the ${i}^{th}$ player across time be $d^{i}_{1}$, $d^{i}_{2}$,$\cdots$, $d^{i}_{n_i}$. The whole range of ranks were divided into three equal segments, with the high, medium and low energy efficient rank groups being $R_{High}$, $R_{Medium}$ and $R_{Low}$ respectively.Let the classes be represented by $C_{sup}^{High}$, $C_{sup}^{Medium}$ and $C_{sup}^{Low}$, where the superscripts signify the energy efficiency behavior of each class. We assign the players to the classes as per the following formula, $P_i$ $\in$ $C_{sup}^{X}$, where,
\begin{align}
    X = \underset{x \in [low,medium, high]}{argmax} \bigg\{\sum_{j=1}^{n_i}\mathds{1}[rank({d^{i}_{j}}) \in R_x]\bigg\}
\end{align}

where $\mathds{1}[\;\cdot\;]$ is the indicator function. This allocates each player into one of the three supervised classes. The behavior of a player in a particular class, e.g. $C_{sup}^{High}$ represents the characteristic behavior of players showcasing high energy efficiency. Then the feature correlations in all the supervised classes and unsupervised clusters were studied using graphical lasso algorithm. Knowledge of feature correlation similarity among members of the supervised classes and unsupervised clusters is used to label the unsupervised clusters  ($C_{unsup}^{1}$, $C_{unsup}^{2}$ and $C_{unsup}^{3}$) as high/medium/low energy efficient. Finally, the labelled unsupervised clusters can be further explored for downstream tasks, such as incentive design. The whole process is illustrated in Figure~\ref{fig:proposed_method}.

\section{Graphical Lasso for Energy Social Game}\label{graph_lasso}
In this section, we formulate a framework towards segmentation analysis that allows us to understand users decision making model. Specifically, we adopt graphical lasso algorithm ~\cite{friedman2008sparse,mainbook} to study the way in which features in a energy game-theoretic framework interplay among each other.

Let the features representing the social game data be denoted by the collection $Y = (Y_1, Y_2, \cdots, Y_S)$. From a graphical perspective, $Y$ can be associated with the vertex set $V =  \{1, 2, \cdots, S\}$ of some underlying graph. The structure of the graph is utilized to derive inferences about the relationship between the features. We use the graphical lasso algorithm ~\cite{mainbook} to realize the underlying graph structure, under the assumption that the distribution of the random variables is Gaussian.

Consider the random variable $Y_s$ at $s$ $\in$ $V$. We use the Neighbourhood-Based Likelihood for graphical representation of multivariate Gaussian random variables. Let the edge set of the graph be given by $\it{E}$ $\subset$ $V\times V$. The neighbourhood set of $Y_s$ is defined by 
\begin{align}
    \mathcal{N}(s) = \{k \in V | (k,s) \in \textit{E}\}
\end{align}
and the collection of all other random variables be represented by:
\begin{align}
    {Y}_{V\backslash\{s\}} = \{Y_{k}, k\in (V-\{s\})\}
\end{align}
For undirected graphical models, node for $Y_s$ is conditionally independent of nodes not directly connected to it given $Y_{\mathcal{N}(s)}$, i.e.
\begin{align}
    (Y_s|{Y}_{V\backslash\{s\}}) \sim (Y_s|{Y}_{\mathcal{N}(s)})
\end{align}
The problem of constructing the inherent graph out of observations is nothing but finding the edge set for every node. This problem becomes predicting the value of $Y_s$ given $Y_{\mathcal{N}(s)}$, or equivalently, predicting the value of $Y_s$ given $Y_{\backslash\{s\}}$by the conditional independence property. The conditional distribution of $Y_s$ given $Y_{\backslash\{s\}}$ is also Gaussian, so the best predictor for $Y_s$ can be written as:
\begin{align}
    {Y}_{s} = {Y}_{V\backslash s}^T.{\beta}^{s} + {W}_{V\backslash s}
\end{align}
where $W_{V\backslash s}$ is zero-mean gaussian prediction error. The $\beta^{s}$ terms dictate the edge set for node s in the graph. We use $l_1$-regularized likelihood methods for getting a sparse $\beta^{s}$. Let the total number of data samples available be \textit{N}. The optimization problem is formulated as: corresponding to each vertex $s = 1, 2, \cdots ,S$, solve the following lasso problem:
\begin{align}
    \hat{{\beta}^{s}} \in \underset{\beta^s \in {\mathbb{R}}^{S-1}}{\rm argmin} \bigg\{{\frac{1}{2N}\sum_{j=1}^{N}(y_{js}-{y^T_{j,{V\backslash s}}}}\beta^{s})^2 + \lambda{\|\beta^s\|}_{1}\bigg\}
\end{align} 
The implementation of Graphical Lasso algorithm is summarized in Appendix.\\

\section{Results}\label{results}
As has been introduced in Section~\ref{sec:method}, we learn the feature correlations using graphical lasso algorithm in supervised classes $C_{sup}^{High}$, $C_{sup}^{Medium}$ and $C_{sup}^{Low}$ to obtain the knowledge about factors governing human decision making towards various (high/medium/low) energy efficient behaviors.  
\newpage
\begin{figure*}
\centering
    { \centering                             
        \stackunder[5pt]\subfigure{\includegraphics[height=0.7cm]{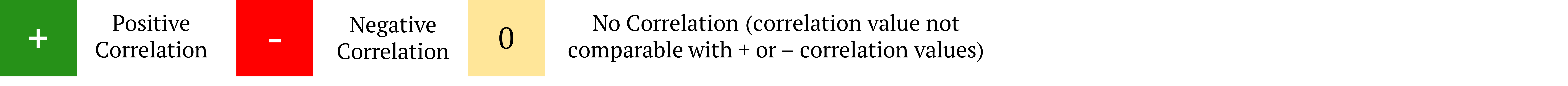}}
    }
    { \centering                             
        \subfigure[]{\includegraphics[width=0.4\textwidth]{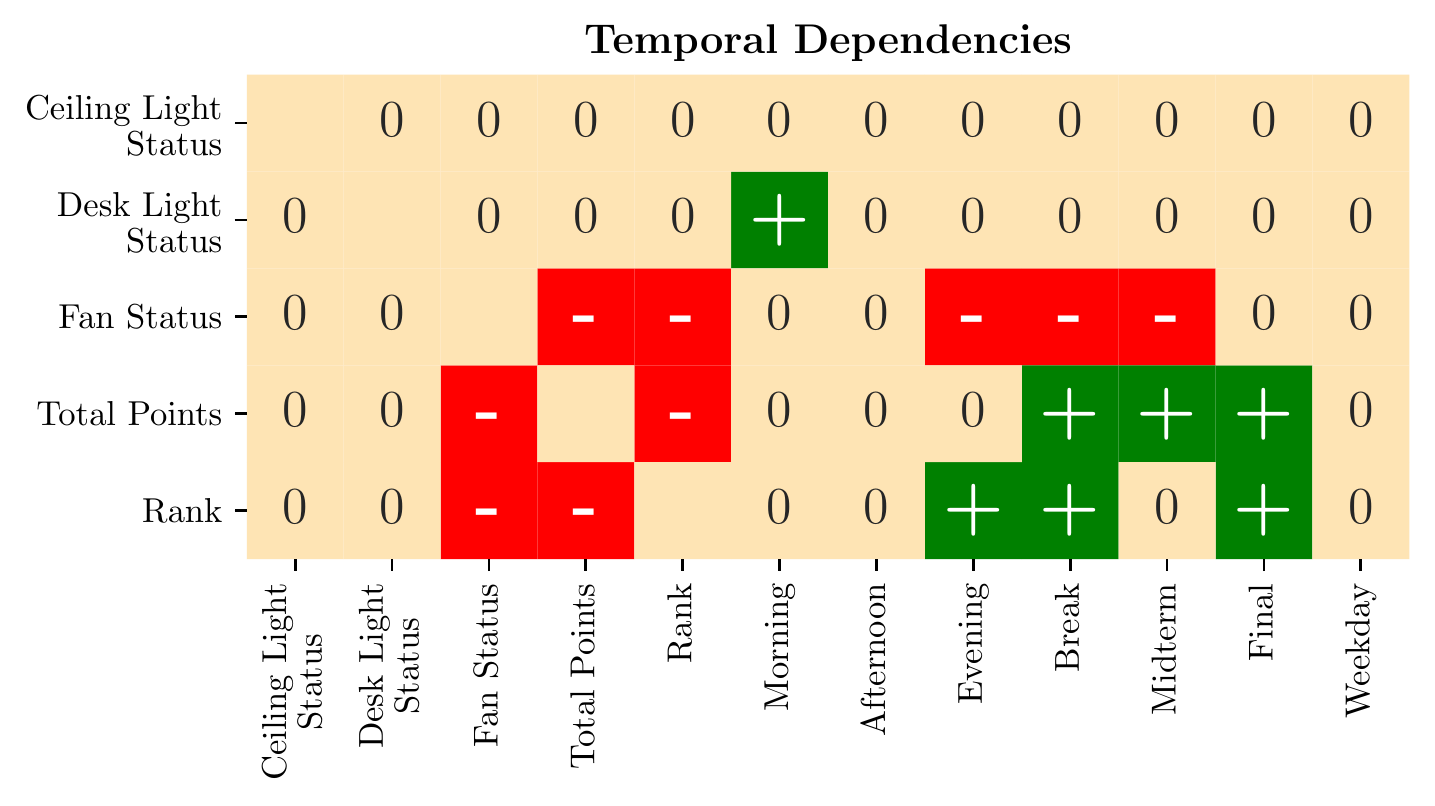}}
        \subfigure[]{\includegraphics[width=0.33\textwidth]{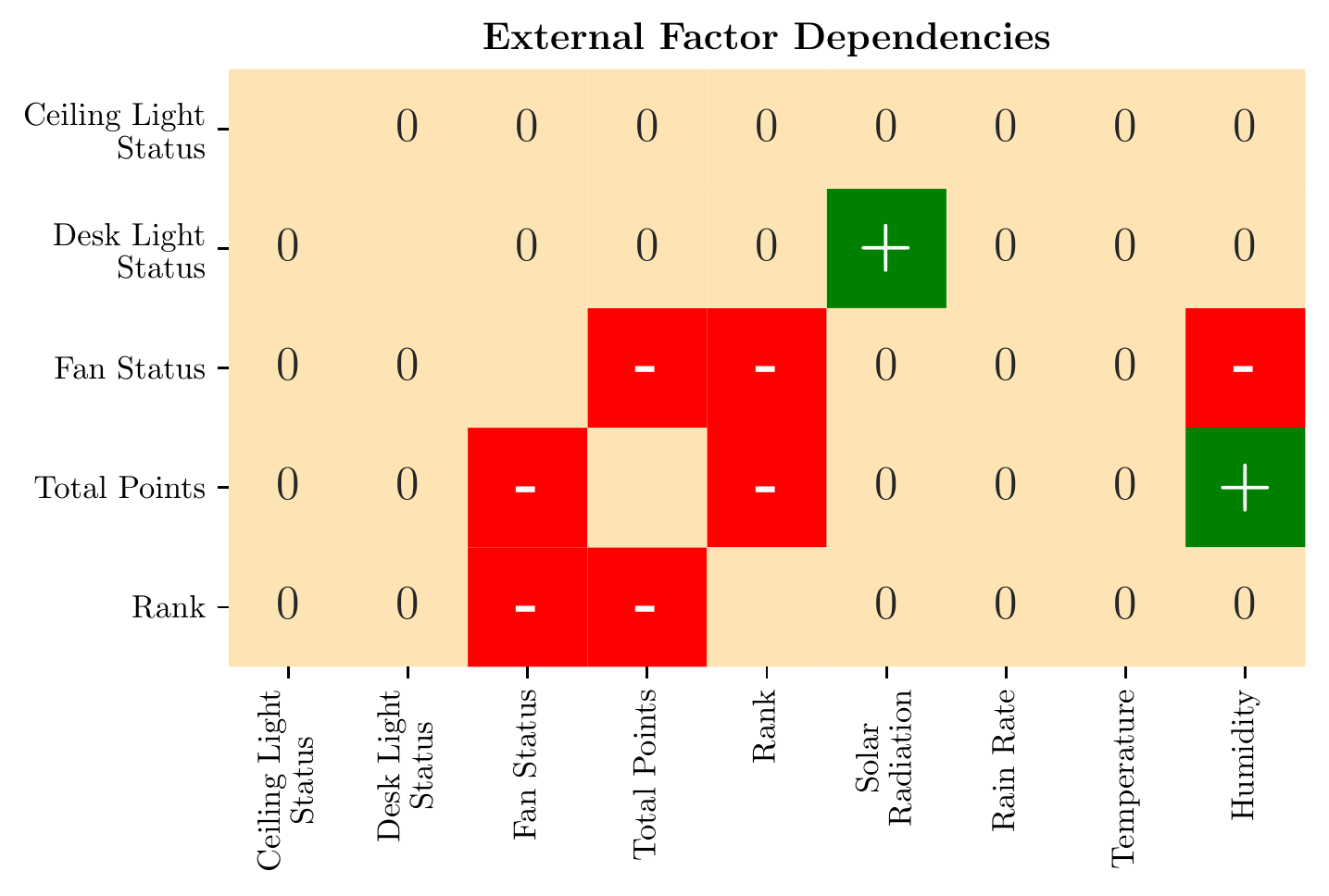}}
        \subfigure[]{\includegraphics[width=0.25\textwidth]{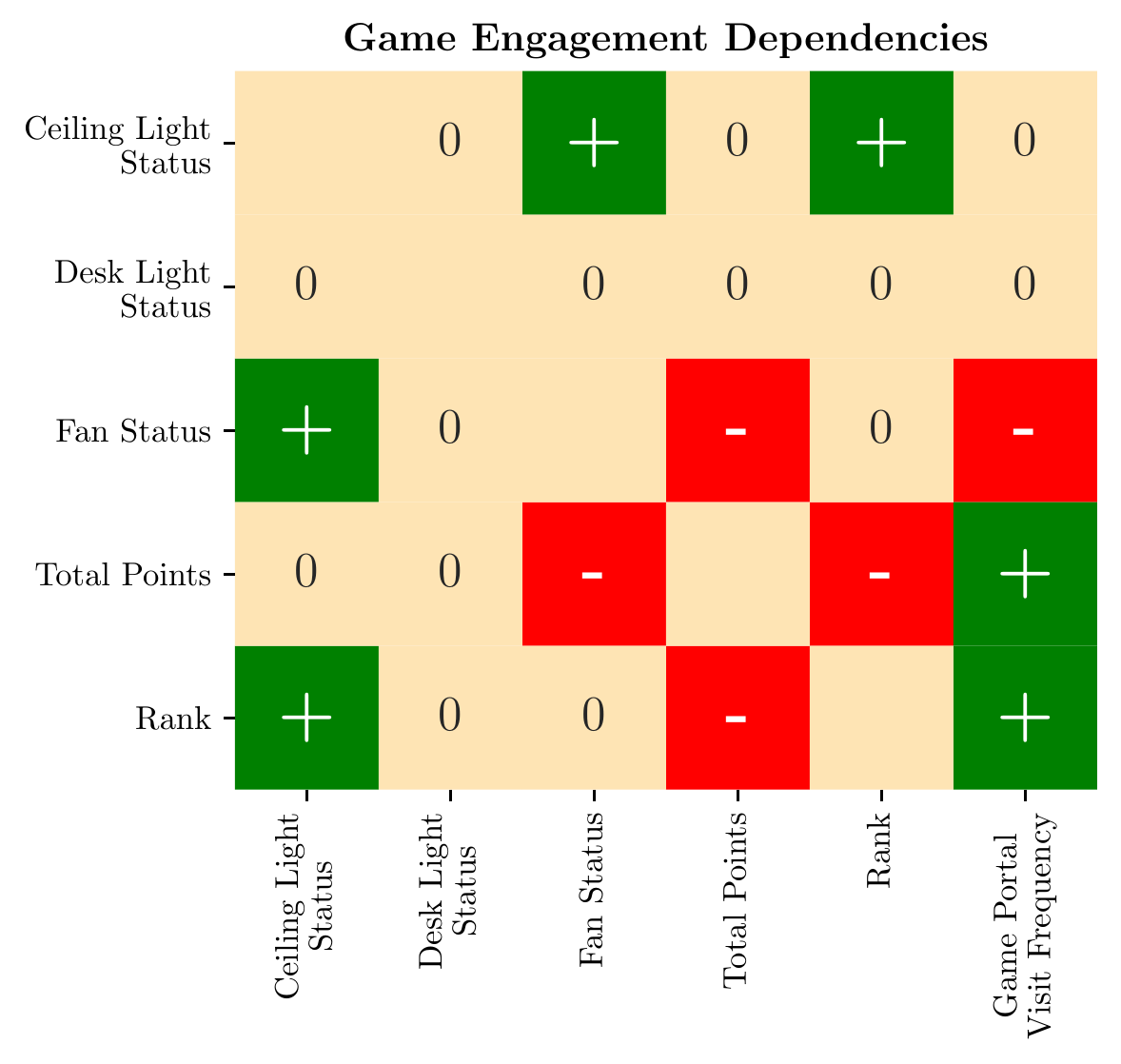}}
    }
    \caption{{\color{black}Feature correlations for a Low Energy Efficient Player ($\in$ $C_{sup}^{Low}$)}}\label{fig_low}
    
    { \centering                             
        \subfigure[]{\includegraphics[width=0.4\textwidth]{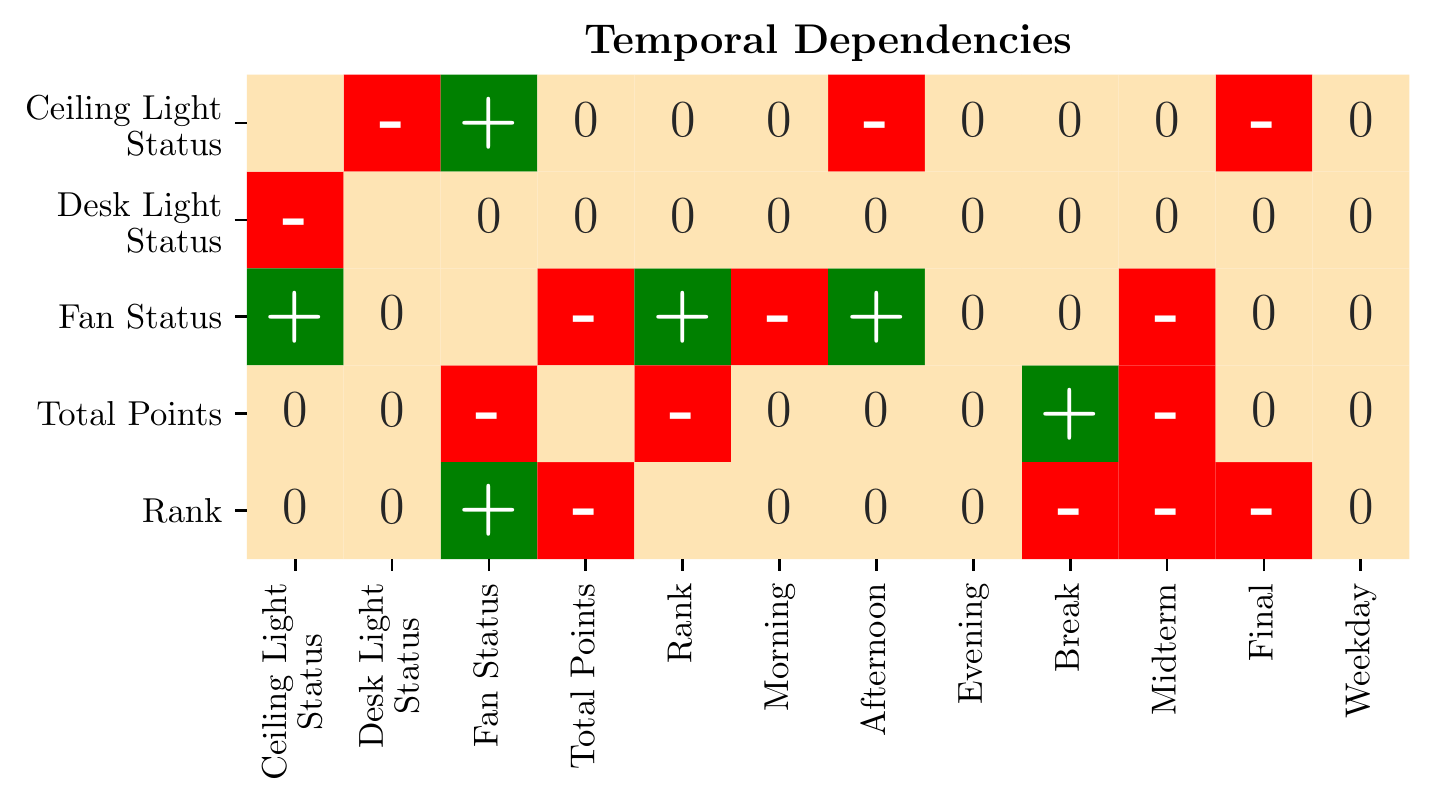}}
        \subfigure[]{\includegraphics[width=0.33\textwidth]{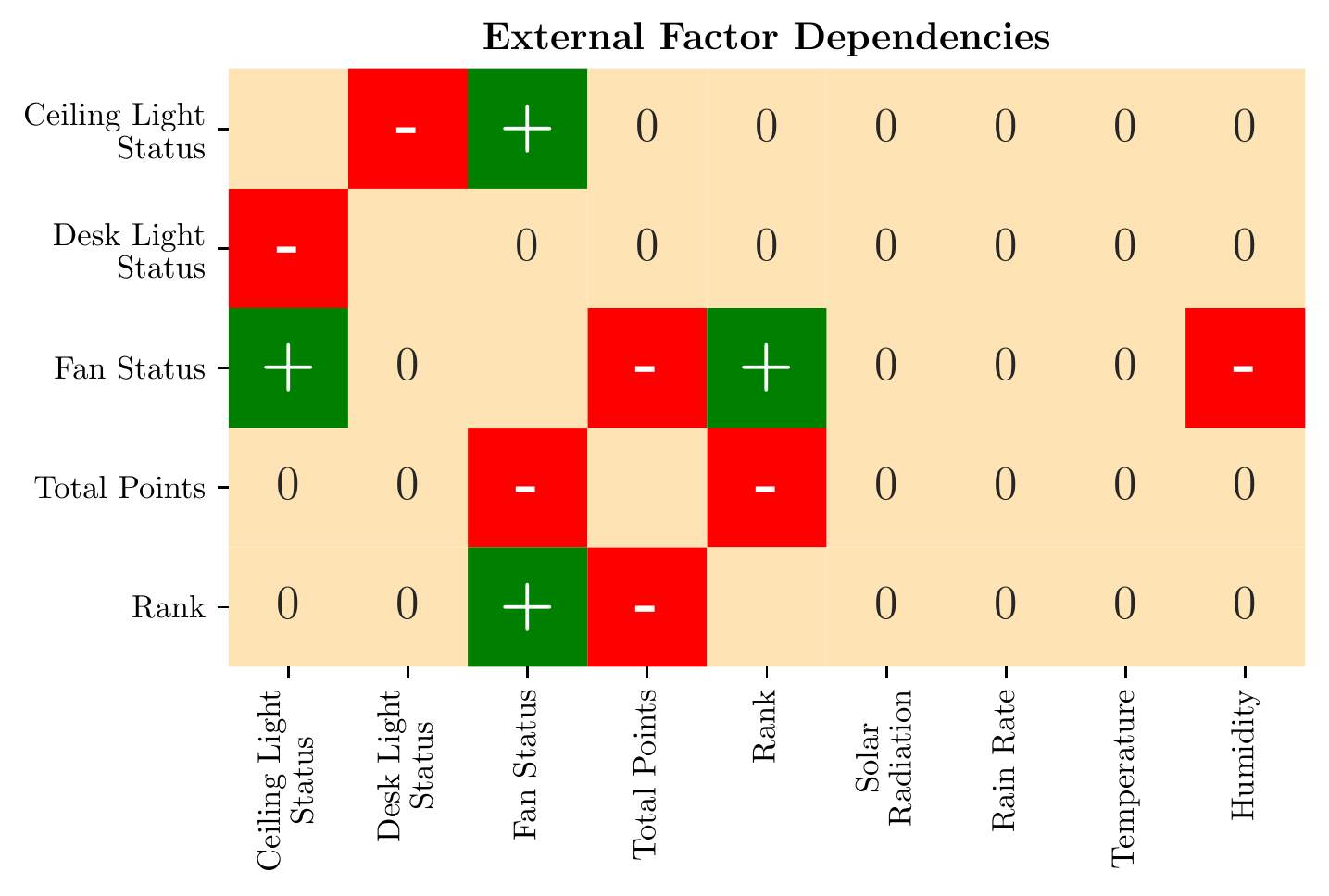}}
        \subfigure[]{\includegraphics[width=0.25\textwidth]{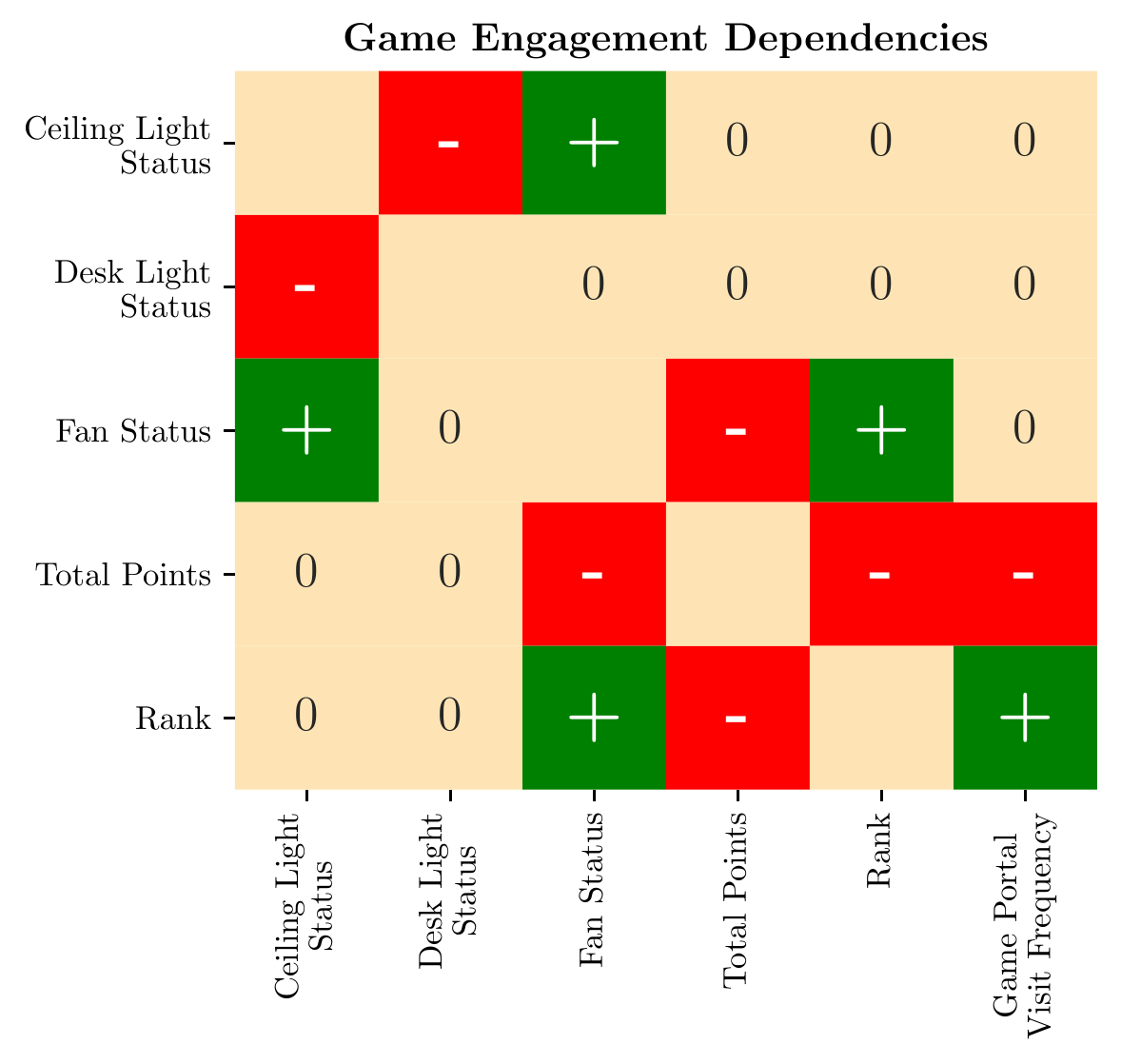}}
    }
    %         \hspace{1cm}
    \caption{{\color{black}Feature correlations for a Medium Energy Efficient Player ($\in$ $C_{sup}^{Medium}$)}}\label{fig_med}
    
    { \centering                             
        \subfigure[]{\includegraphics[width=0.4\textwidth]{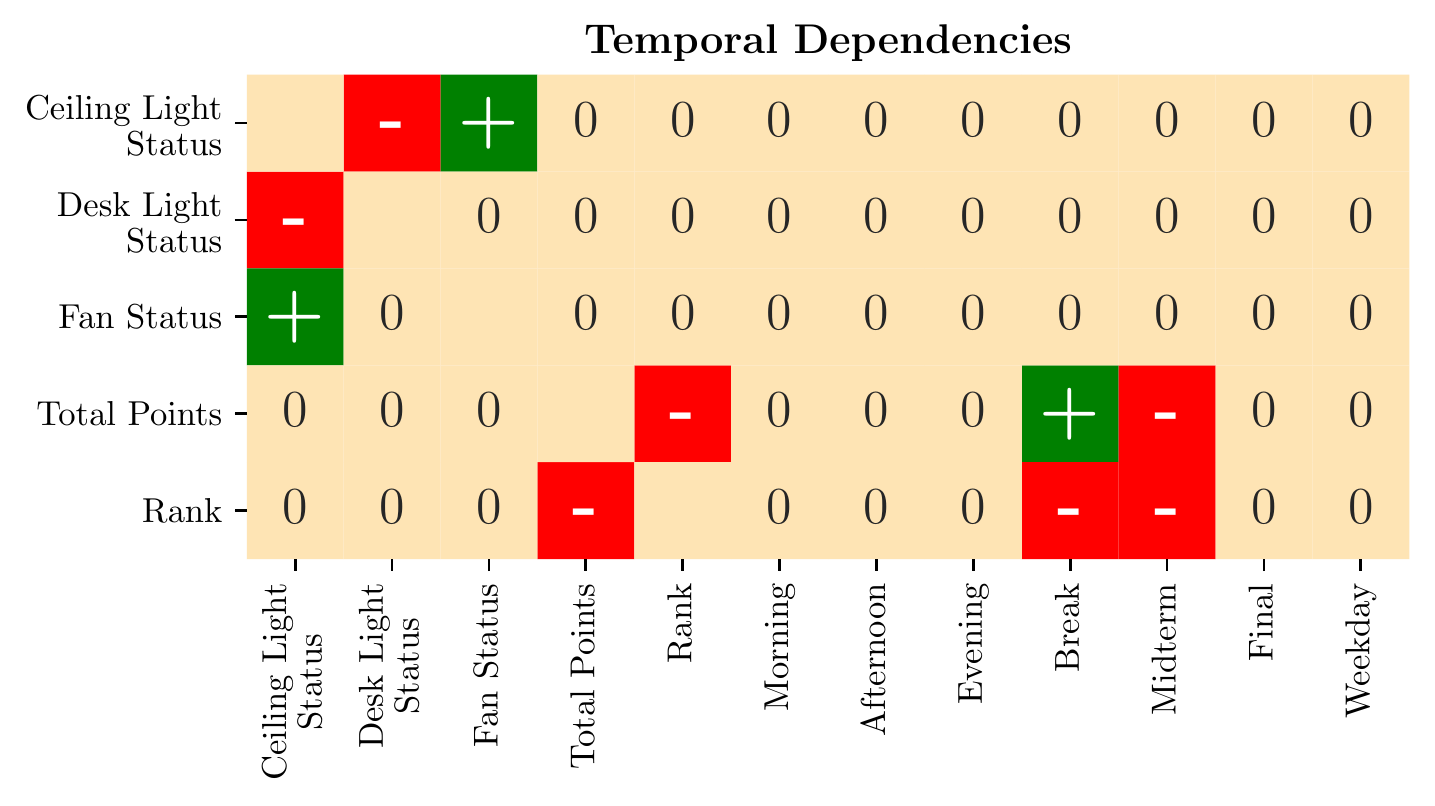}}
        \subfigure[]{\includegraphics[width=0.33\textwidth]{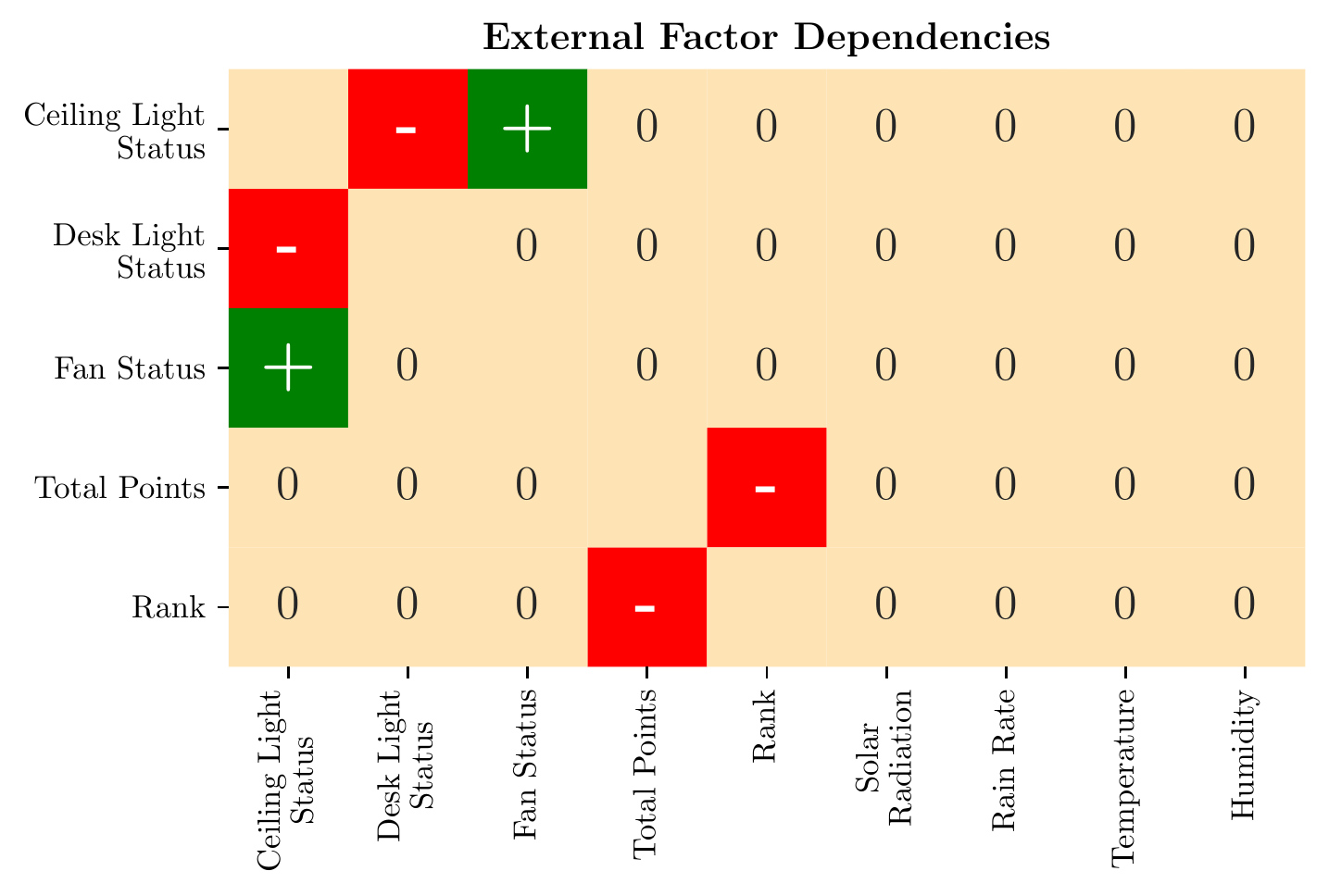}}
        \subfigure[]{\includegraphics[width=0.25\textwidth]{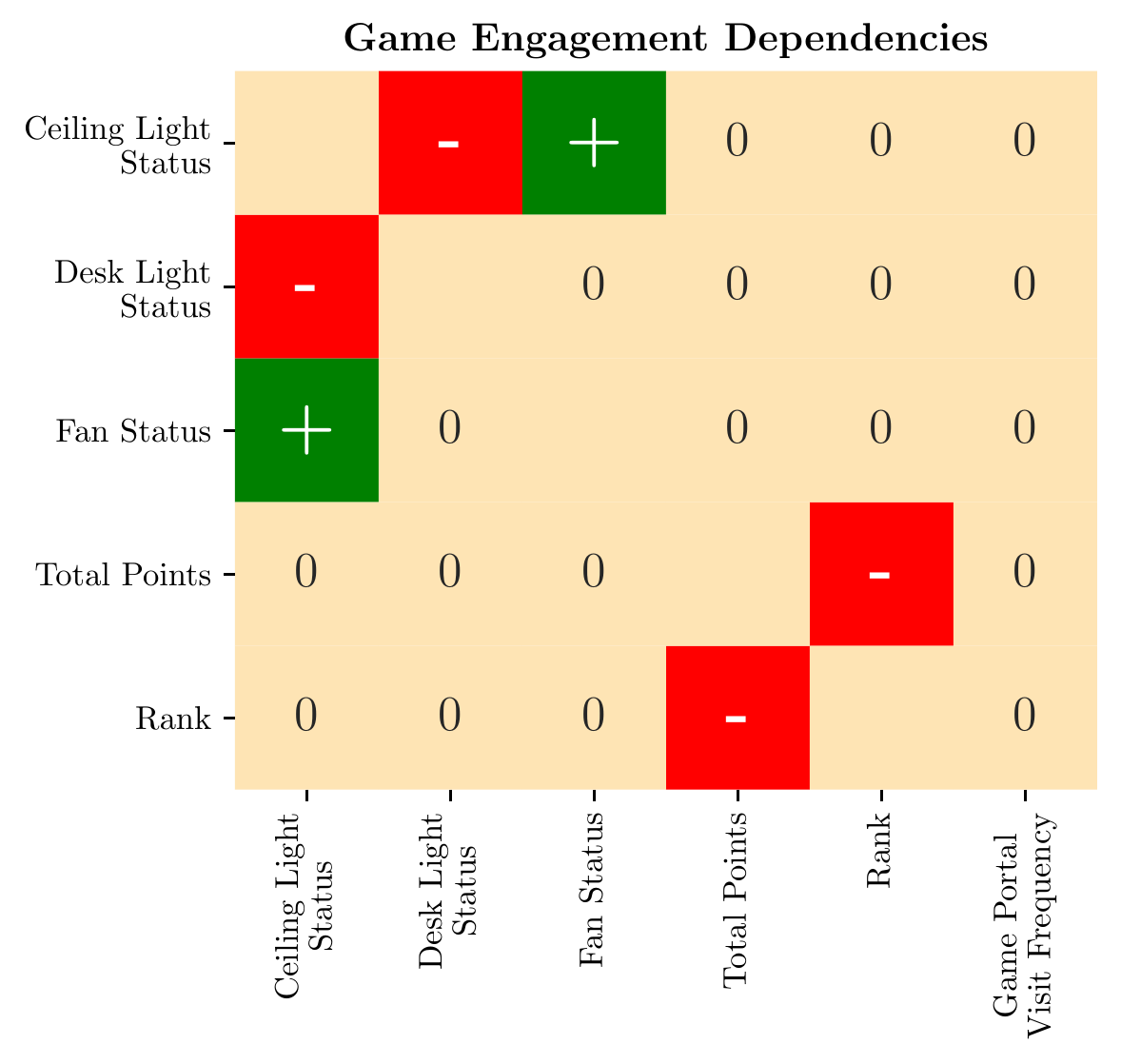}}
    }
    %         \hspace{1cm}
    \caption{{\color{black}Feature correlations for a High Energy Efficient Player ($\in$ $C_{sup}^{High}$)}}\label{fig_high}
    
    { \centering                             
        \subfigure[]{\includegraphics[width=0.4\textwidth]{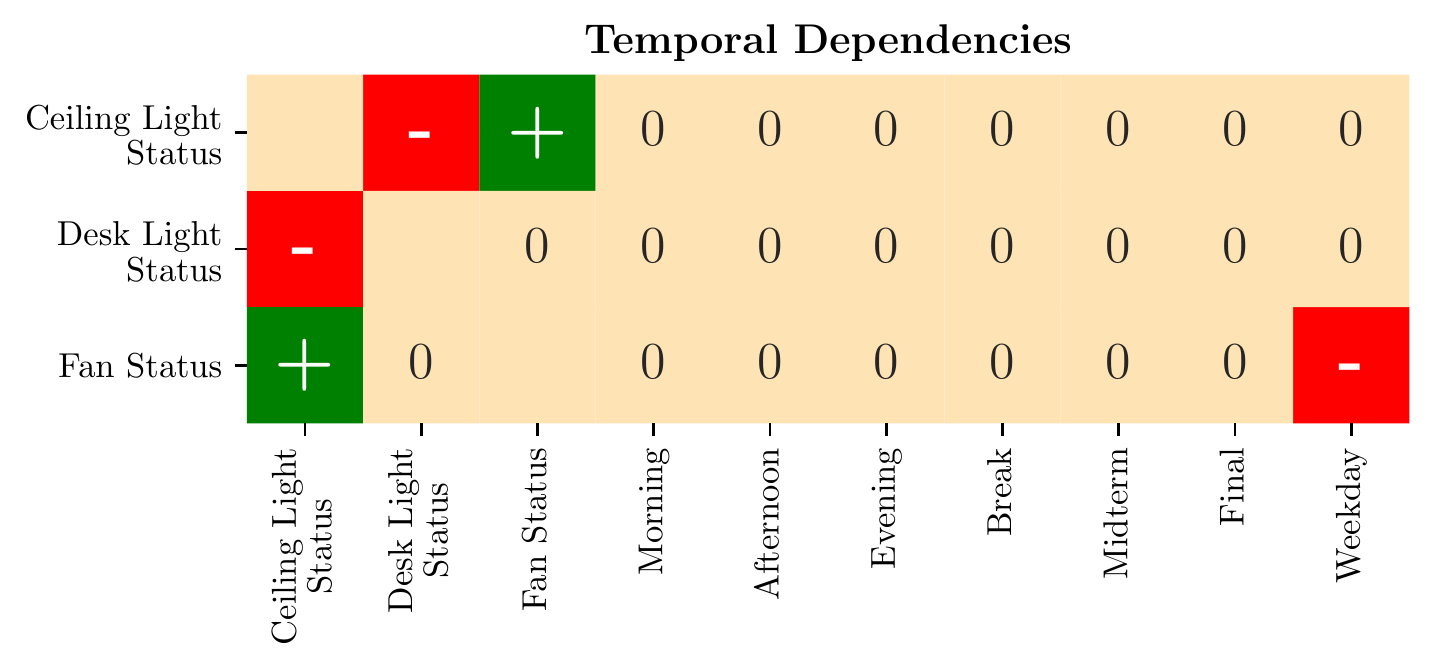}}
        \subfigure[]{\includegraphics[width=0.33\textwidth]{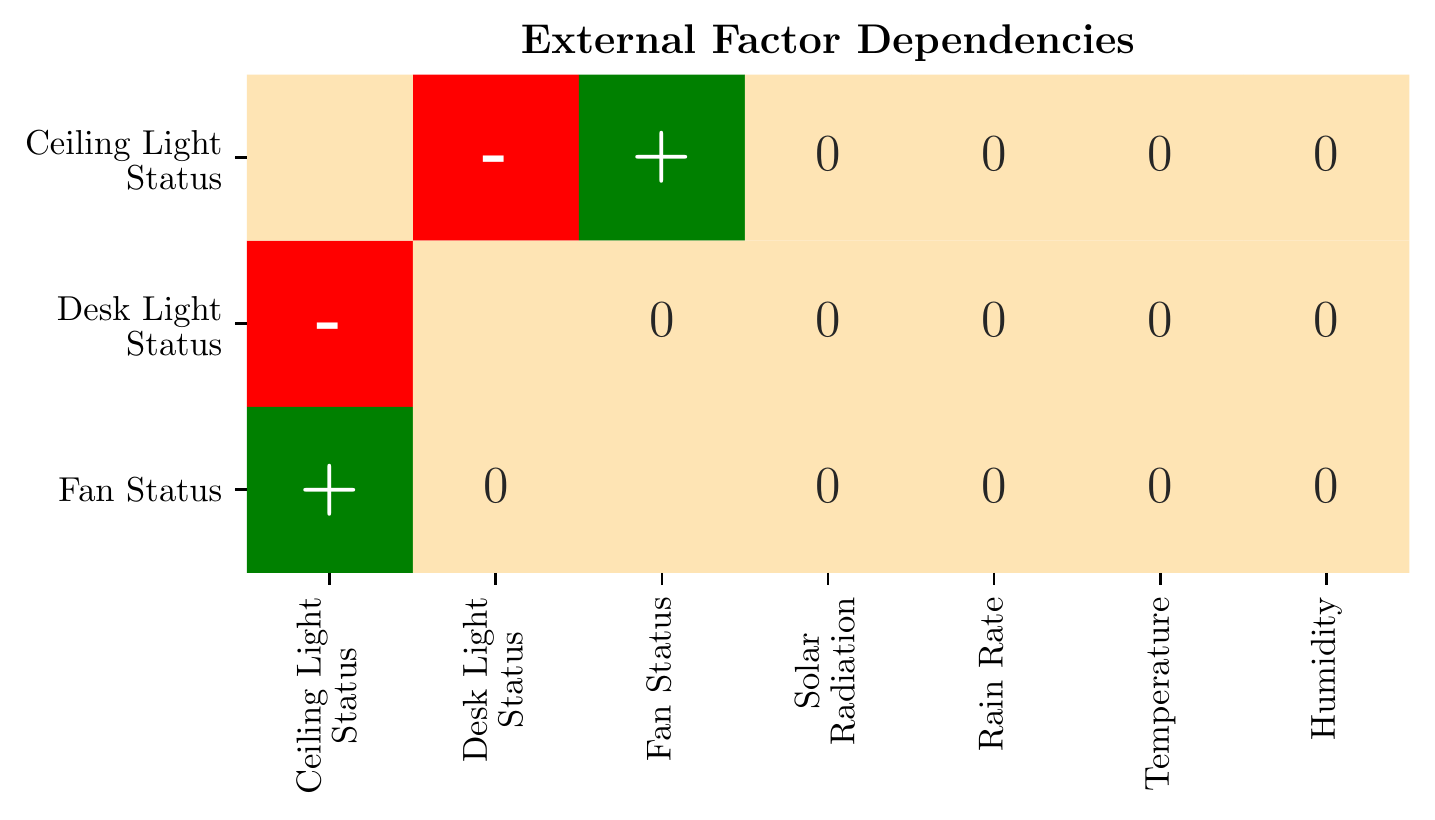}}
        \subfigure[]{\includegraphics[width=0.25\textwidth]{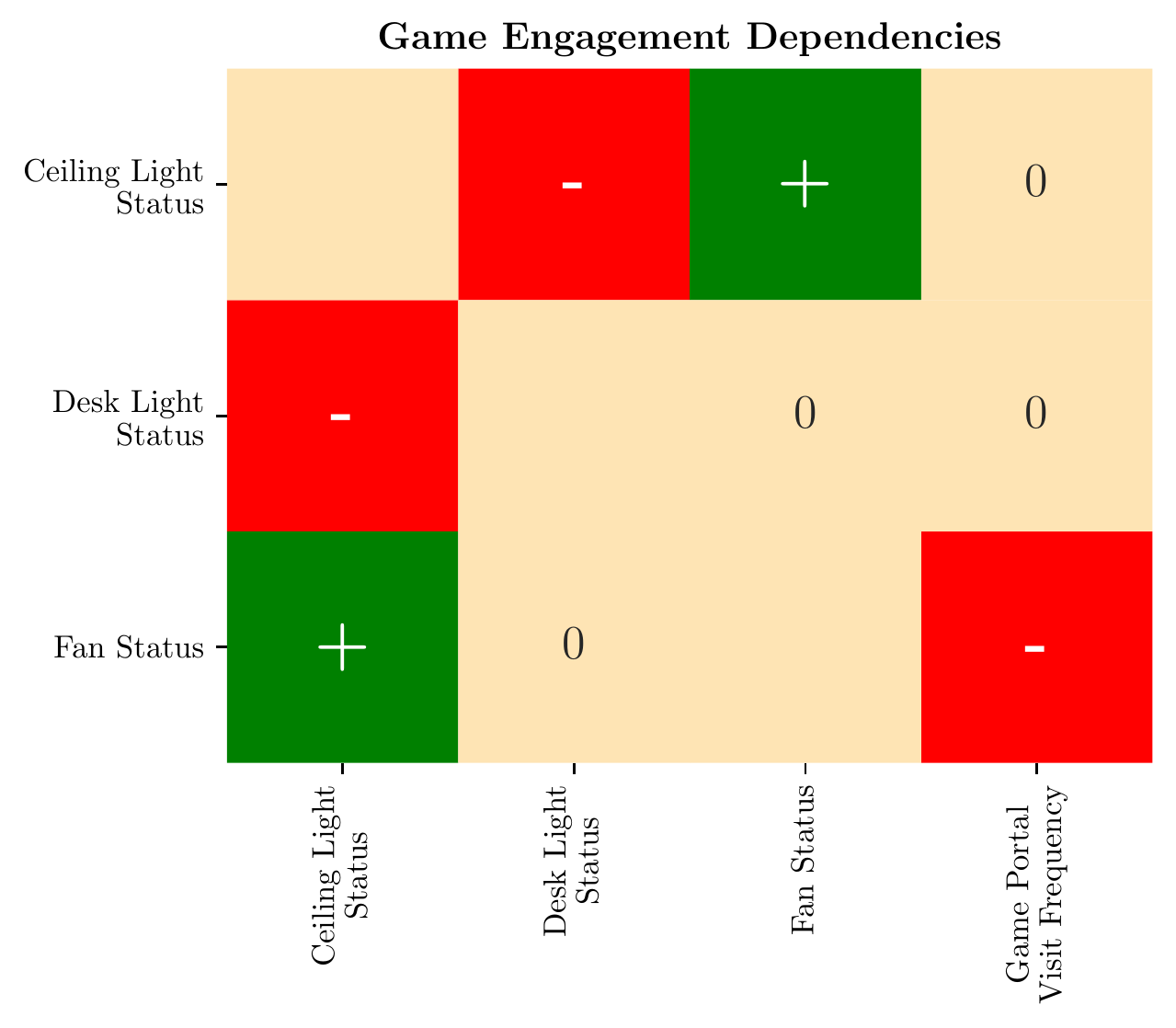}}
    }
    %         \hspace{1cm}
    \caption{{\color{black}Feature correlations for energy usage behaviors in $C_{unsup}^{3}$. The labels "Total Points" and "Rank" are removed for unsupervised clustering.}}\label{fig_cluster3}
\end{figure*}
\clearpage
\subsection{Feature correlation learning in supervised segregation}\label{sec:glasso_supervised}
We consider a representative player (selected as the player holding the median rank in the class) for each of the three classes obtained out of supervised segregation method described in Section~\ref{sec:method} to run graphical lasso and study the correlation between the features for that class. We group the features into different categories so as to study their influence on energy efficiency behaviors. Specifically, we consider \emph{Temporal} features like time of the day, academic schedules and weekday/weekends, \emph{External} features as outdoor temperature, humidity, rain rate etc. and \emph{Game Engagement} features like frequency of visits to game web portal. 

The feature correlations for a low energy efficient player is given in Fig~\ref{fig_low}. The player tries to use each resource independently which can be observed in Figure~\ref{fig_low}(a) with no correlation between the corresponding resource usage identifiers. There is a positive correlation between morning and desk light usage indicating heedless behavior towards energy savings. The absolute energy savings increase during the breaks and finals, given by positive correlation with total points, but it is not significant as compared to other players during the same period, thus increasing the rank. External parameters play a significant role in energy usage behavior of this class. The operation of the ceiling fan is driven by external humidity as given in Figure~\ref{fig_low}(b). Figure~\ref{fig_low}(c) indicates that their frequency of visits to the game web portal is motivated by sub-optimal performance in the game.

Feature correlations for a medium energy efficient player is given in Fig~\ref{fig_med}. The player showcases predictable behaviors with correlations between desk light, ceiling light and ceiling fan usage (Figure~\ref{fig_med}(a)). The player co-optimizes the usage by alternating the use of ceiling and desk light. Different occasions like break, midterm and final are marked by energy saving patterns. Unlike a low energy efficient player, the player in this class tries to save energy in a conscious manner shown by reduced fan usage during the morning and reduced light usage during the afternoon. The fan usage is influenced by the external humidity, shown by Fig~\ref{fig_med}(b). The game engagement patterns for a player in this class (Fig~\ref{fig_med}(c)) is similar to that of the low energy efficient class.

Fig~\ref{fig_high} shows the feature correlations for a high energy efficient player. This player also exhibits predictable behavior. Opportunistically, this player saves energy during breaks and midterms as shown by negative correlation between the corresponding flags and rank in Figure~\ref{fig_high}(a). Notice that there exists a negative correlation between midterm flag and total points, indicating decrease in absolute amount of points. However, the points are still higher than the points by other players which marks improvement in the rank. This behavior is completely opposite to what is  exhibited by a player in low energy efficient class. The player is neither affected by the time of the day, nor by the external factors (Figure~\ref{fig_high}(b)) showing a dedicated effort to save energy. The game engagement behavior for this player, given in Figure~\ref{fig_high}(c) is inconclusive, possibly due to dominance by other energy saving factors.
\begin{table*}[!ht]
  \centering
  \setlength\arrayrulewidth{0.8pt}
  \resizebox{2.05\columnwidth}{!}{
  \begin{tabular}{|c|c|c|c|c|c|c|c|c|c|c|c|c|c|c|c|}
  \rowcolor{Gray}
  \hline
    \multicolumn{1}{|c|}{Test whether $X$ causes $Y$} & \multicolumn{2}{c|}{Fan $\Rightarrow$ Ceiling Light} & \multicolumn{2}{c|}{Humidity $\Rightarrow$ Fan} & \multicolumn{2}{c|}{Desk Light $\Rightarrow$ Fan} & \multicolumn{2}{c|}{Ceiling Light $\Rightarrow$ Desk Light} & \multicolumn{2}{c|}{Morning $\Rightarrow$ Desk Light} & \multicolumn{2}{c|}{Afternoon $\Rightarrow$ Fan} & \multicolumn{2}{c|}{Evening $\Rightarrow$ Ceiling Light} \\ \hline
    \rowcolor{Gray}
Player type & p-value & F-statistic & p-value & F-statistic & p-value & F-statistic & p-value & F-statistic & p-value & F-statistic & p-value & F-statistic & p-value & F-statistic\\ \hline
Low Energy Efficient & 0.54 & 0.37 & \cellcolor{blue!25}\textbf{0.004} & 8.12 & 0.06 & 3.55 & 0.81 & 0.06 & 0.4 & 0.71 & \cellcolor{blue!25}\textbf{0.01} & 6.1 & \cellcolor{blue!25}\textbf{0} & 25.3\\ \hline
Medium Energy Efficient & \cellcolor{blue!25}\textbf{0} & 21.2 & \cellcolor{blue!25}\textbf{0.008} & 7.06 & \cellcolor{blue!25}\textbf{0} & 113.6 & \cellcolor{blue!25}\textbf{0} & 25.8 & 0.23 & 1.41 & 0.46 & 0.55 & \cellcolor{blue!25}\textbf{0.0007} & 11.5\\ \hline
High Energy Efficient & \cellcolor{blue!25}\textbf{0} & 21.9 & 0.12 & 2.36 & 0.99 & 0.003 & 0.93 & 0.007 & 0.63 & 0.22 & \cellcolor{blue!25}\textbf{0.04} & 4.2 & 0.52 & 0.41\\ \hline
\end{tabular}}
\caption{Causality test results among various potential causal relationships. In bold are the p-values (shaded in blue) in cases that Granger causality is established through F-statistic test between features. p-values lower than $0.05$ indicate strong causal relationship in 5\% significance level}
\label{tab:grangers_causality}
\end{table*}

\subsection{Causal Relationship between features}
To ensure the correctness of results in Section~\ref{sec:glasso_supervised} and to enhance the explainable nature of our model, we studied the causal relationship between features using Granger causality test. Granger causality is a statistical test used to determine causal relationship between two signals. If signal A granger-causes signal B, then past values of A can be used to predict B for future timesteps beyond what is available for B. The results for causal relationship study is given in Table~\ref{tab:grangers_causality}. Under null hypothesis $H_0$, $X$ does not Granger-cause $Y$. So, a p-value lower than $0.05$ (5\% significance level) indicates a strong causal relationship between the tested features and implies rejecting the null hypothesis $H_0$.

The p-values (shaded in blue) for which Granger causality is established are highlighted in the table. Interestingly, for medium and high energy efficient building occupants, ceiling fan usage causes ceiling light usage. This in fact confirms the predictive behavior for them as mentioned earlier. In both low and medium energy efficient building occupants, external humidity causes ceiling fan usage. This is an indicator that their energy usage is affected by external weather conditions. However, for high energy efficient building occupants external humidity doesn't cause ceiling fan usage. This shows that they are highly engaged with the proposed gamification interface and try to minimize their energy usage. Another interesting result is that the evening label causes ceiling light usage for both low and medium energy efficient building occupants. But this is not the case for high energy efficient building occupants, for whom ceiling light usage is better optimized as a result of their strong engagement with the ongoing social game, eventually leading to exhibition of better energy efficiency.
% \subsection{Labelling of Unsupervised Clusters using Feature Correlations Knowledge of Supervised Classes}
\subsection{Labelling unsupervised clusters using feature correlation knowledge from supervised classification}
We also learn the feature correlations in clusters obtained from unsupervised clustering of data in Section~\ref{sec:method}. Based on the feature correlation knowledge gained from different supervised classes in Section~\ref{sec:glasso_supervised}, we label the clusters as having low, medium or high energy efficient data. As an illustration, the feature correlations for $C_{unsup}^3$ is shown in Fig~\ref{fig_cluster3}. It is evident from Fig~\ref{fig_cluster3}(a) that data in $C_{unsup}^3$ exhibit predictability in behavior with correlations between resource usage flags. Also the weekdays are marked by energy savings in terms of decrease in fan usage minutes. The time of the day is also unrelated to the performance. Neither do the external factors contribute to the performance (Figure~\ref{fig_cluster3}(b)). The engagement in the game also boosts the points (Figure~\ref{fig_cluster3}(c)). All the above behaviors are indicative of the similarity between the energy efficiency characteristics manifested by $C_{unsup}^3$ and the high energy efficient class obtained using supervised segregation ($C_{sup}^{High}$). So, $C_{unsup}^3$ is labelled as the high energy efficient cluster. Following the same comparison, $C_{unsup}^1$ and $C_{unsup}^2$ are labelled as the medium and low energy efficient clusters respectively. 

\begin{figure}[!ht]
\centering
\includegraphics[width=0.45\textwidth]{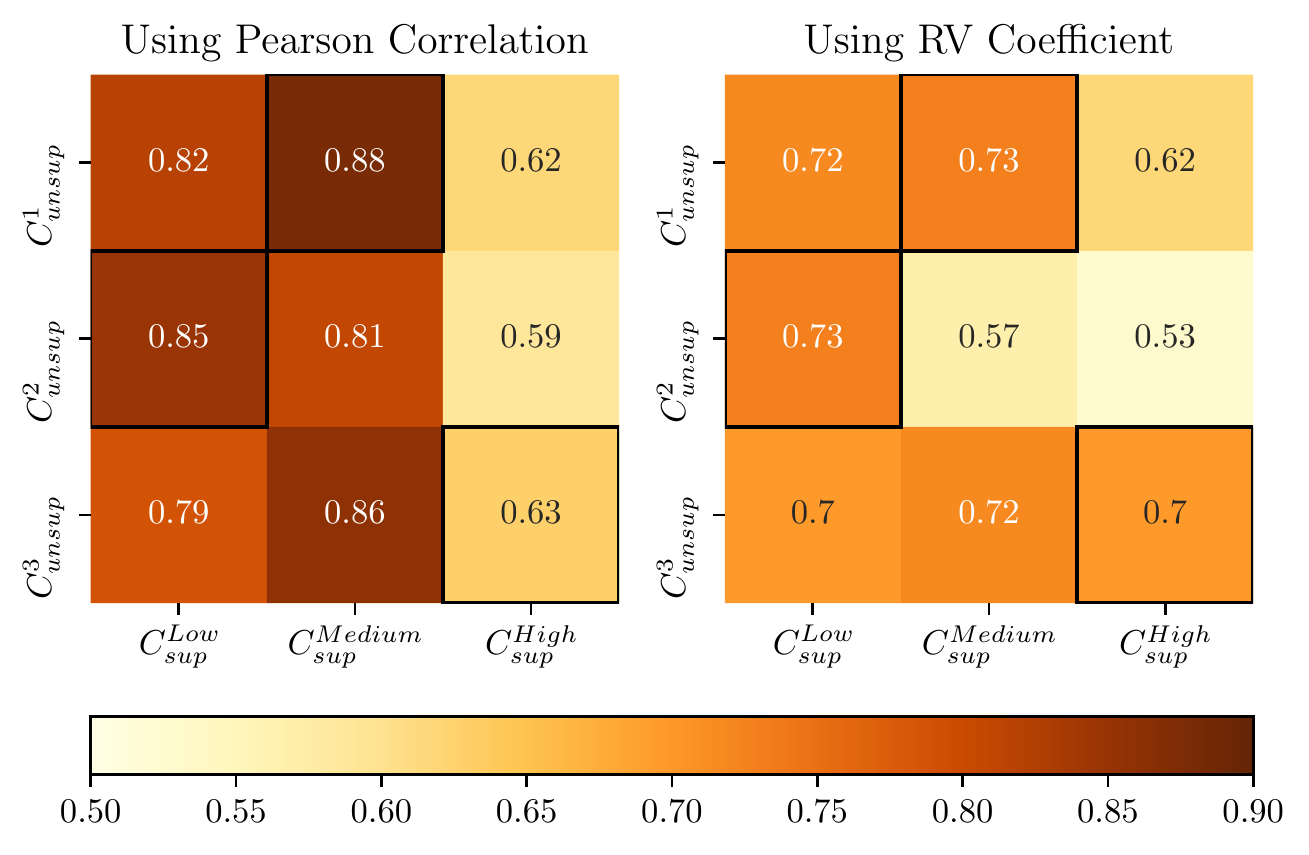}
\caption{Similarity between feature correlation matrices. The highest value in each column is highlighted and corresponds to the matching of supervised classes to the unsupervised clusters}
\label{fig:pearson_corr}
\end{figure}

To further strengthen our inference, we compute the similarity using Pearson Correlation and RV coefficient~\cite{robert1976a} between the feature correlation matrices in unsupervised clusters and supervised classes. Figure~\ref{fig:pearson_corr} showing the result of above operation confirms our earlier assignment of labels to the unsupervised clusters, i.e. \{$C_{unsup}^1$ $\sim$ Medium Energy Efficient\},\{$C_{unsup}^2$ $\sim$ Low Energy Efficient\} and \{$C_{unsup}^3$ $\sim$ High Energy Efficient\}.  The labelled unsupervised clusters are the final segments that can be used for a number of downstream tasks as discussed in Section~\ref{sec:conclusion}.

\section{Conclusion and Future Work}\label{sec:conclusion}
A general framework for segmentation analysis in energy game-theoretic frameworks was presented in this research work. The analysis included clustering of agent behaviors and an explainable statistical model representing the contributed features motivating their decision-making. To strengthen our results, we examined several feature correlations using granger causality test for potential causal relationships. Coupled with the statistical justification and explainable nature, the proposed method can provide characteristic clusters demonstrating different energy usage behaviors, following which, specific incentives can be designed for each cluster.

\begin{figure}[h]
\centering
\includegraphics[width=0.48\textwidth]{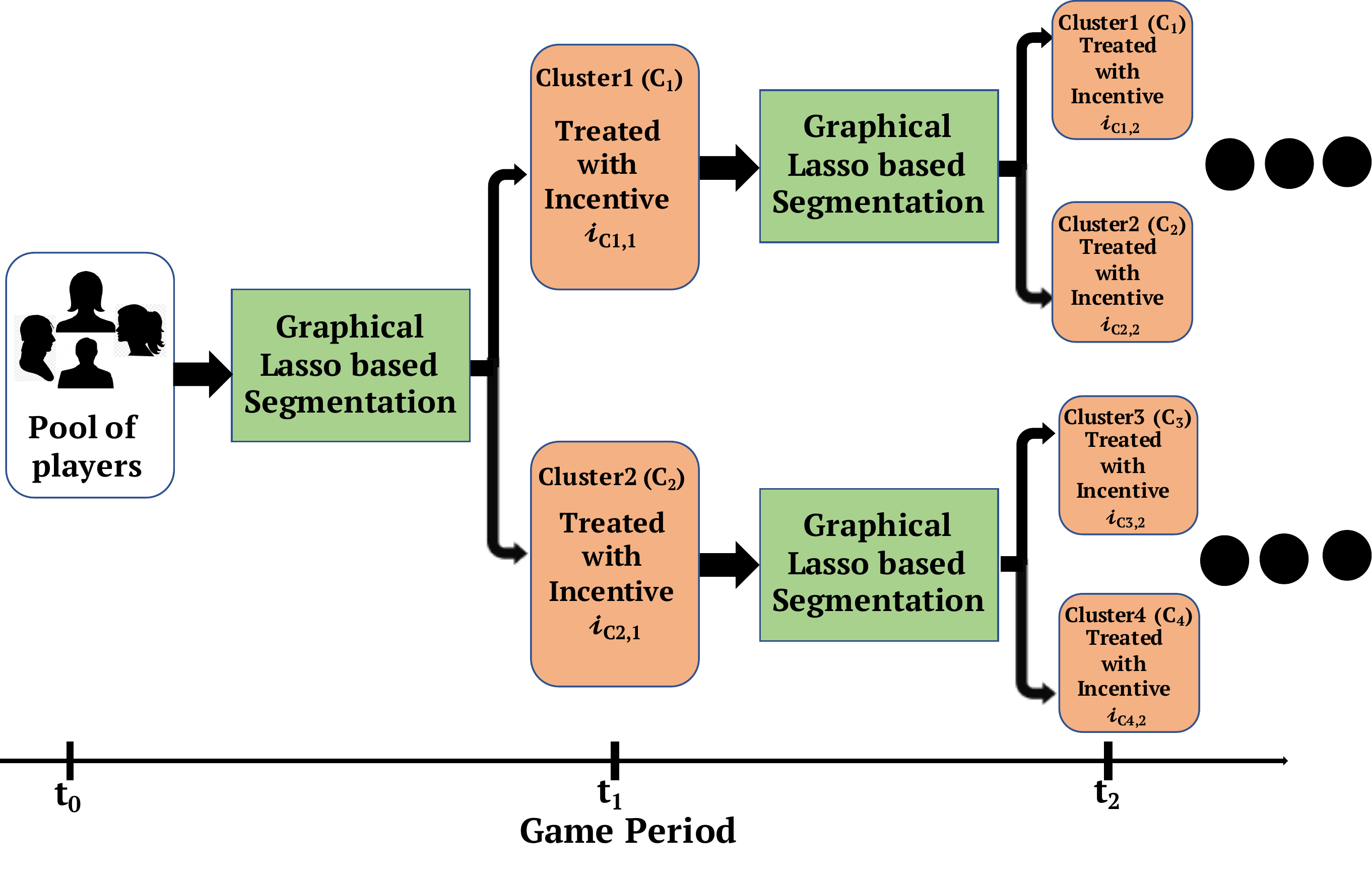}
\caption{Tree based incentive design mechanism employing proposed graphical lasso based segmentation method. Clusters are treated with incentives specifically tailored for them.}
\label{fig:iterative_game}
\end{figure}

There are several directions for future research. Our ultimate goal for the segmentation analysis is to improve the gamification methodology, to simultaneously learn occupant preferences while also opening avenues for feedback, as static programs for encouraging energy efficiency are less efficient with passing of time~\cite{schipper:2000aa}. So, an improved version of energy social game, similar in structure to that of ~\cite{konstantakopoulos2019deep} but with intelligent incentive design and privacy preserving techniques~\cite{jia2018poisoning} can be implemented, with building occupants and managers interaction modeled as a reverse stackelberg game (leader-follower) in which there are multiple followers that play in a  non-cooperative game setting~\cite{ratliff2014social}. By leveraging proposed segmentation analysis, an adaptive model can be formulated that learns how user preferences change over time, and thus generate the appropriate incentives. Furthermore, the learned preferences can be adjusted through incentive mechanisms~\cite{ratliff2018adaptive} and a tailored mean-field game approach~\cite{gomes2018mean} to enact improved energy efficiency. Above two operations can be carried out in a tree structure, with segmentation carried out in regular intervals in each of the tree branches, as depicted in Figure ~\ref{fig:iterative_game}. This can be coherently designed with other smart building systems~\cite{zou2019machine,zou2019wifi,liu2019personal,biscuit}. Summing up, this would result in a novel mechanism design, effectively enabling variation in occupant's behaviors, in order to meet, for instance, the requirements of a demand response program. Another line of future work can be to study the delayed impacts of energy social game and design it accordingly to achieve long term energy efficiency, like a research in same line~\cite{liu2018delayed}.
\section{Acknowledgments}
This research is funded by the Republic of Singapore's National Research Foundation through a grant to the Berkeley Education Alliance for Research in Singapore (BEARS) for the Singapore-Berkeley Building Efficiency and Sustainability in the Tropics (SinBerBEST) Program. BEARS has been established by the University of California, Berkeley as a center for intellectual excellence in research and education in Singapore. The work of I. C. Konstantakopoulos was supported by a scholarship of the Alexander S. Onassis Public Benefit Foundation.
\bibliographystyle{unsrt}
\small{
\bibliography{glasso}}
\section*{Appendix}
\noindent \textbf{A. Description of Graphical Lasso algorithm}\\
\begin{tikzpicture}[yscale=3] 
\draw [line width=0.65mm, black ] (0,-1) -- (8.5,-1) node [right]{};;
\end{tikzpicture}
 Algorithm 1: GRAPHICAL LASSO ALGORITHM FOR\\GAUSSIAN GRAPHICAL MODELS\\
\begin{tikzpicture}[yscale=3] 
\draw [line width=0.65mm, black ] (0,0) -- (8.5,0) node [right]{};;
\end{tikzpicture}
\begin{enumerate}[1.]
\item For vertices $s = 1, 2, \cdots ,S$:
%1. For vertices s = 1,2,....,F:
\begin{enumerate}[a.]
    \item Calculate initial loss ${\|Y_{s} - Y^T_{V\backslash s}\beta^{s}\|}^{2}_{2}$
    \item Untill Convergence:
    \begin{enumerate}[i.]
        \item Calculate partial residual $r^{(s)}$ = $Y_s$ - $Y^T_{V\backslash s}\beta^{s}$
        \item For all $j \in {V\backslash s}$, Get ${\beta}^{s,new}_j$ = ${\textbf S_{\lambda}\big(\frac{1}{N}\langle r^{(s)},Y_j\rangle\big)}$ 
        \item Compute new loss = ${\|Y_{s} - Y^T_{V\backslash s}\beta^{s,new}\|}^{2}_{2}$
        \item Update ${\beta^{s}}$ = ${\beta}^{s,new}$
    \end{enumerate}
    \item Get the neighbourhood set $\mathcal{N}$(s)= supp(${{\beta}}^{s}$) for s
\end{enumerate}
\item Combine the neighbourhood estimates to form a graph estimate ${G} = (V, \textit{E}$) of the random variables.
\end{enumerate}
\begin{tikzpicture}[yscale=3] 
\draw [line width=0.65mm, black ] (0,-1) -- (8.5,-1) node [right]{};;
\end{tikzpicture}

$\textbf S_{\lambda}(\theta)$ is soft thresholding operator as $sign(\theta)(|\theta|-\lambda)_{+}$.\\
\hspace{50mm}

For optimal design of penalty factor $\lambda$ in Graphical Lasso run for a vertex s, we take 10 values in logarithmic scale between $\lambda_{max}$ and $\lambda_{min}$ as given below and conduct a line search to find the penalty factor which brings the minimum loss.
\begin{align}
 \lambda_{max} = \frac{1}{N}\underset{j\in V\backslash s}{max}|\langle Y_j,Y_s\rangle|, \quad \lambda_{min} = \frac{\lambda_{max}}{100}
 \end{align}
Implementing a coordinate descent approach ~\cite{mainbook}, the time complexity of the proposed algorithm is $O(SN)$ for a complete run through all $S$ features. We also do 5-fold cross validation to ensure accurate value of the coefficients $\beta^{s}$. Use of partial residuals for each node significantly reduces the time complexity of the algorithm.
\end{document}